\documentclass[conference,onecolumn]{IEEEtran}
\IEEEoverridecommandlockouts
\usepackage{cite}
\usepackage{multirow}
\usepackage{amsmath,amssymb,amsfonts}
\usepackage{algorithmic}
\usepackage{graphicx}
\usepackage{textcomp}
\usepackage{xcolor}
\newcommand{\RNum}[1]{\uppercase\expandafter{\romannumeral #1\relax}}

\def\BibTeX{{\rm B\kern-.05em{\sc i\kern-.025em b}\kern-.08em
    T\kern-.1667em\lower.7ex\hbox{E}\kern-.125emX}}
\begin{document}
\title{Clustering Enabled Few-Shot Load Forecasting}

\author{\IEEEauthorblockN{Qiyuan Wang$^*$, Zhihui Chen$^*$}
\IEEEauthorblockA{\textit{School of Data Science}\\
\textit{The Chinese University of Hongkong, Shenzhen} \\
Shenzhen, Guangdong, 518172 China \\
\{qiyuanwang, zhihuichen\}@link.cuhk.edu.cn}
\and
\IEEEauthorblockN{Chenye Wu$^\dag$}\
\IEEEauthorblockA{\textit{School of Science and Engineering}\\
\textit{The Chinese University of Hongkong, Shenzhen} \\
Shenzhen, Guangdong, 518172 China \\
chenyewu@yeah.net}

}

\maketitle

\renewcommand{\thefootnote}{\fnsymbol{footnote}}
\footnotetext[1]{The first two authors contributed equally to this work, and hence are co-first authors of this work.}
\footnotetext[2]{C. Wu is the correspondence author. This work was supported in part by Shenzhen Institute of Artificial Intelligence and Robotics for Society.}

\begin{abstract}
While the advanced machine learning algorithms are effective in load forecasting, they often suffer from low data utilization, and hence their superior performance relies on massive datasets. This motivates us to design machine learning algorithms with improved data utilization. Specifically, we consider the load forecasting for a new user in the system by observing only few shots (data points) of its energy consumption. This task is challenging since the limited samples are insufficient to exploit the temporal characteristics, essential for load forecasting. Nonetheless, we notice that there are not too many temporal characteristics for residential loads due to the limited kinds of human lifestyle. Hence, we propose to utilize the historical load profile data from existing users to conduct effective clustering, which mitigates the challenges brought by the limited samples. Specifically, we first design a feature extraction clustering method for categorizing historical data. Then, inheriting the prior knowledge from the clustering results, we propose a two-phase Long Short Term Memory (LSTM) model to conduct load forecasting for new users. The proposed method outperforms the traditional LSTM model, especially when the training sample size fails to cover a whole period (i.e., 24 hours in our task). Extensive case studies on two real-world datasets and one synthetic dataset verify the effectiveness and efficiency of our method.
\end{abstract}


\section{Introduction}
Load forecasting is an essential tool in the energy sector, and it is the basis for decision-making in power system control and electricity market operation. Over the past decade, the demand for more accurate forecasts enlarged the application of deep learning techniques in load forecasting, together with the need for training with large-scale data. Load forecasting techniques based on recursive neural network (RNN) \cite{coulibaly2005nonstationary}, and feature learning models like convolutional neural network (CNN) \cite{borovykh2017conditional} achieve good performance by extracting complex statistics and learning representative features from data. However, such models often rely on large-scale load data for training \cite{wang2020generalizing}, i.e., deep learning suffers from poor sample efficiency, in stark contrast to classical time series and regression approaches. As shown in Fig. 1, when only provided with a limited number of samples with supervised information (shots), the prediction results of deep learning models such as Long Short Term Memory (LSTM) are far from satisfactory. 
%Load forecasting, a classical procedure in the electricity sector, is the basis for efficient power system control and effective electricity market operation. Over the past two decades, the rising deep learning technologies have fundamentally reshaped the toolbox for load forecasting. For example, dependency learning models such as recursive neural network (RNN) \cite{coulibaly2005nonstationary}, and feature learning models like convolutional neural network (CNN)\cite{borovykh2017conditional} are both capable of extracting complex statistics and learning representative features from massive datasets (Change the references). However, such models are often not very data-efficient, i.e., deep learning approaches suffer from poor sample efficiency in stark contrast to [human perception]. As shown in Fig. 1, when only provided with a [short] sequence of historical data, the prediction results of deep learning models such as Long Short Term Memory (LSTM) are far from satisfactory.

\begin{figure}[htbp]
    \centering
    \includegraphics[width=0.5\linewidth,scale=1.0]{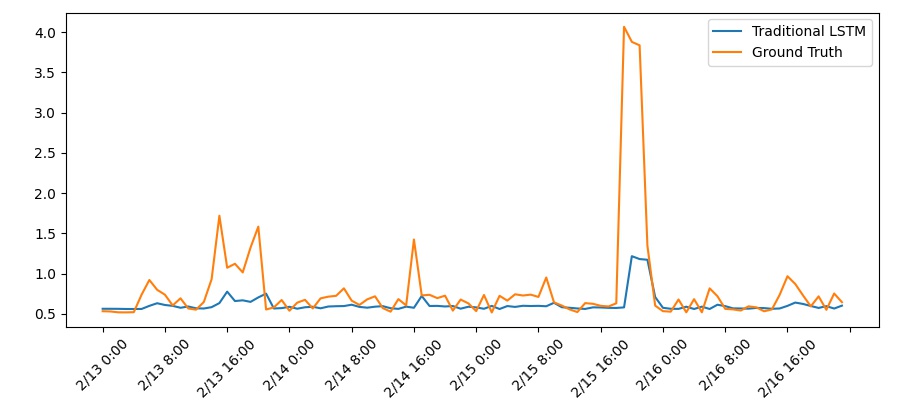}
    \caption{96 Hours LSTM Forecasting on 12-shot Training Set}
    \label{Figure1}
\end{figure}

\begin{figure*}[htbp]
    \centering
    \includegraphics[width=0.5\linewidth,scale=0.500]{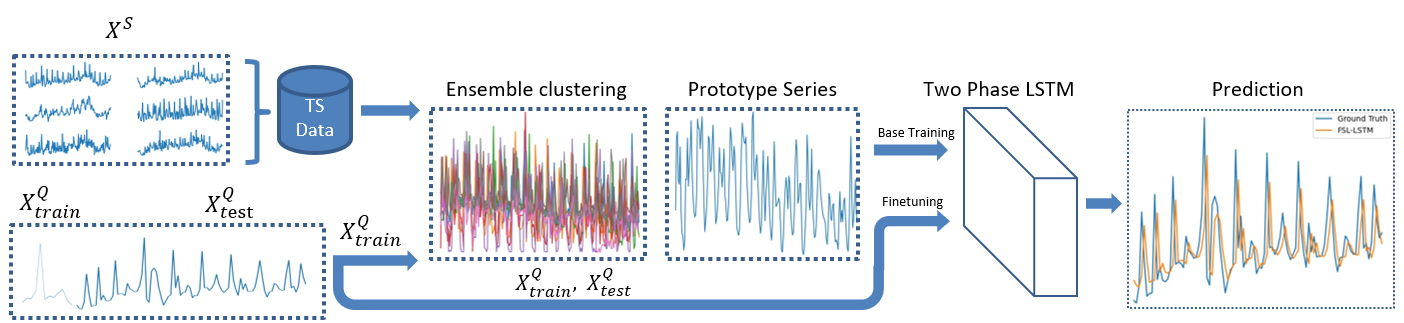}
    \caption{The Framework of FSL-LSTM}
    \label{Figure2}
\end{figure*}
%Commonly recognized by the community, deep learning models'  innate dependence on large scale training data can be hard to weaken \cite{wang2020generalizing, folmsbee2018active, he2020sample}. 
Reducing deep learning models' innate dependence on large-scale training data and obtaining good load forecasting performance when limited target data is available can be challenging. The intuition of our work is to utilize the patterns in load profiles to help the forecasting model improve its performance when training data for the target users are limited. Specifically, there is a finite number of underlying daily energy consumption patterns \cite{van1999cluster}. Thus, given a short sequence of energy consumption profiles from a target user (an unknown sample), we may identify its consumption pattern. Once such identification is successful, the load forecasting model may utilize rich historical data from the given consumption pattern as a source of prior knowledge and the short sequence of the target user as fine-tuning data. A collective training strategy is then introduced to enable good forecasting performance on the target user with limited data available.

%While Such an observation is not surprising. we may utilize the patterns in load profiles to help LSTM improve its performance. Specifically, there are limited number of underlying daily energy consumption patterns \cite{van1999cluster}. Thus, we may first observe a short sequence of energy consumption profile from a new user (an unknown sample) and try to identify its consumption pattern. Once such identification is successful, the LSTM may utilize information in the pattern as rich historical data and the short sequence as short term memory to improve the performance. We term this task the clustering enabled few-shot load forecasting.

Specifically, this work integrates ensemble clustering and a two-phase LSTM model to achieve better forecasting accuracy based on few-shot samples. As shown in Fig. 2, limited samples from the target user (few-shot samples) will first be classified into similar clusters with rich historical data from numerous users. Then the LSTM model will first be trained (this can be done offline and hence being a pretrained model) with the long-term denoised mean-averaging data from the specific cluster. By further fine-tuning the pretrained model with the few-shot samples, the resulting two-phase LSTM can utilize the prior knowledge of the cluster and the real-time information of the target user. Together, the proposed model achieves a remarkable performance under the shortage of training data for the target user.

The remainder of the paper is organized as follows. Section \RNum{2} reviews the literature on time series forecasting, clustering, and few-shot learning (FSL). Then, Section \RNum{3} introduces our proposed two-phase LSTM model in detail. To validate the performance of few-shot forecasting, we introduce the performance metrics, datasets overview, and case study design in Section \RNum{4}. Comprehensive numerical studies are conducted in Section \RNum{5}. Finally, Section \RNum{6} gives the concluding remarks and points out interesting future directions.

\section{Related works}
We identify three major streams of related works. The first one seeks to apply time series forecasting in the electricity sector. The second one investigates the time series clustering techniques, while the third one targets to advance FSL.

\subsection{Time Series Forecasting in Electricity Sector}
Time series forecasting is applied in the electricity sector to facilitate decision-making. Notably, in the electricity sector, load forecasting has long been an important research topic. Statistical and machine learning-based methods are widely applied in load forecasting. In \cite{huang2003short}, Huang and Shih presented an Auto-regressive moving average (ARMA) procedure for load forecasting characterizing the non-Gaussian process. The ARMA model can be extended to Auto-regressive Integrated Moving Average (ARIMA) model, which is widely used in forecasting electricity load and market price\cite{contreras2003arima}. 

Recently, machine learning techniques have become popular in load forecasting. In \cite{park1991electric}, Park $et$ $al.$ presented a multi-layered perceptron artificial neural network (ANN) that interpolates among the load and temperature data.
In \cite{dcd123}, Elman neural network-based forecast engine with empirical mode decomposition was proposed as a novel method for predicting load signal. Introduced by Hochreiter $et$ $al$. in \cite{6795963}, LSTM has received enormous attention in this area due to its capacity of capturing long-distance statistical regularities, e.g., in \cite{8039509,7885096,probLSTM}, LSTM-based deep learning forecasting frameworks were used in load forecasting.

\subsection{Time Series Clustering}
Time series clustering has been a hot topic in data mining. Compared with the classical clustering method, time series clustering is more complicated due to the temporal dynamics. Therefore, compared to the standard clustering methods, time series clustering also cares about similarity measurement and feature extraction.

The most classical time series clustering is based on temporal similarity metrics, such as Euclidean distance (ED) and dynamic time warping (DTW). Although such distance metrics are easy to implement in practice, they suffer from fatal demerits: ED suffers from the dimensionality curse \cite{verleysen2005curse}, while DTW is overly sensitive to local changes.

To overcome the demerits of similarity-based clustering, former researchers have investigated feature extraction-based clustering methods. Such methods first extract the key features in the time series and then cluster in low dimensional feature space. Thus, they can better capture the global feature of time series. The most fundamental feature extraction tools include Discrete Fourier transform (DFT) and discrete cosine transform (DCT). Another widely adopted feature extraction tool is the discrete wavelet transform (DWT). In this work, we follow the novel feature extraction workflow based on DWT in \cite{hacine2018wavelet}, where Hacine-Gharbi $et$ $al$. proposed wavelet cepstral coefficient (WCC) for feature extraction and then utilized a hidden Markov model for electricity appliance identification. This procedure achieves a completeness ratio of $98.13\%$ when the decomposition level is five.

\subsection{Few-Shot Learning}
The objective of FSL is to learn new tasks supported by only a few samples with supervised information. FSL enables the learning of rare cases and relieves the burden of large-scale data collection. One approach is to constrain hypothesis space $\mathcal{H}$ by prior knowledge in the learning process. For example, Caruana proposed Multitask Learning \cite{caruana1997multitask}, an inductive transfer mechanism to improve generalization performance by using domain information contained in training signals of related tasks.  

Another approach is to alter the search strategy in hypothesis space $\mathcal{H}$ by using prior knowledge extracted from a set of relevant tasks to provide a good initialization or guide the search steps. A popular approach is to apply meta-learning to continuously refine the parameters according to the learner's past performance. One representative method is model-agnostic meta-learning (MAML), proposed in \cite{articalMAML}. Also, many efforts have been devoted to achieving FSL by fine-tuning the parameter from a good initialization, including those based on generated-adversarial network (GAN)\cite{Liu_2019_ICCV} and CNN\cite{yan2019meta}. However, to our best knowledge, few attempts have been made to extend these approaches to LSTM for time series forecasting. In our work, we use historical load data of multiple users to provide a good initialization that enables LSTM to adapt to novel load forecasting tasks quickly.

\section{FSL for Load Forecasting}
Our proposed FSL framework consists of two major components: the primary ensemble clustering and a two-phase LSTM forecasting network. We use compact selected features extracted from wavelet analysis and other statistic descriptors for the first component. For the second component, we follow \cite{6795963} to implement the LSTM, utilizing wavelet denoising and model fine-tuning.

\subsection{Feature Extraction for Clustering}\label{AA}
\subsubsection{Discrete wavelet analysis}
The whole procedure starts with an ensemble clustering where few-shot samples are clustered with historical data. The historical data of numerous users are segmented according to the length and the point in time of the few-shot samples ($k$-shot) to ensure the clustering is conducted on the series representing the same period of time. In order to reduce the dimensionality of the sequence set, wavelet analysis is adopted to project the original data onto a lower-dimensional feature space. We compute three descriptors in wavelet analysis: discrete wavelet energy (DWE), log wavelet energy (LWE), and WCC.

As proposed in \cite{hacine2018wavelet}, we follow a feature extraction workflow based on wavelet analysis (shown in Fig. 3). However, instead of applying DWT where merely the low frequency coefficients are decomposed, we use discrete wavelet packet transform (DWPT) to decompose both low and high frequency components at each stage for a more comprehensive feature abstraction. By applying DWPT to the historical load data, an original time series will be converted into a balanced tree structure. In each level $j$, the total number of wavelet samples is equal to ${2}^{j}$, where each leaf node represents a set of wavelet coefficients either in high or low frequency.

\begin{figure}[htbp]
    \centering
    \includegraphics[width=0.5\linewidth,scale=1.00]{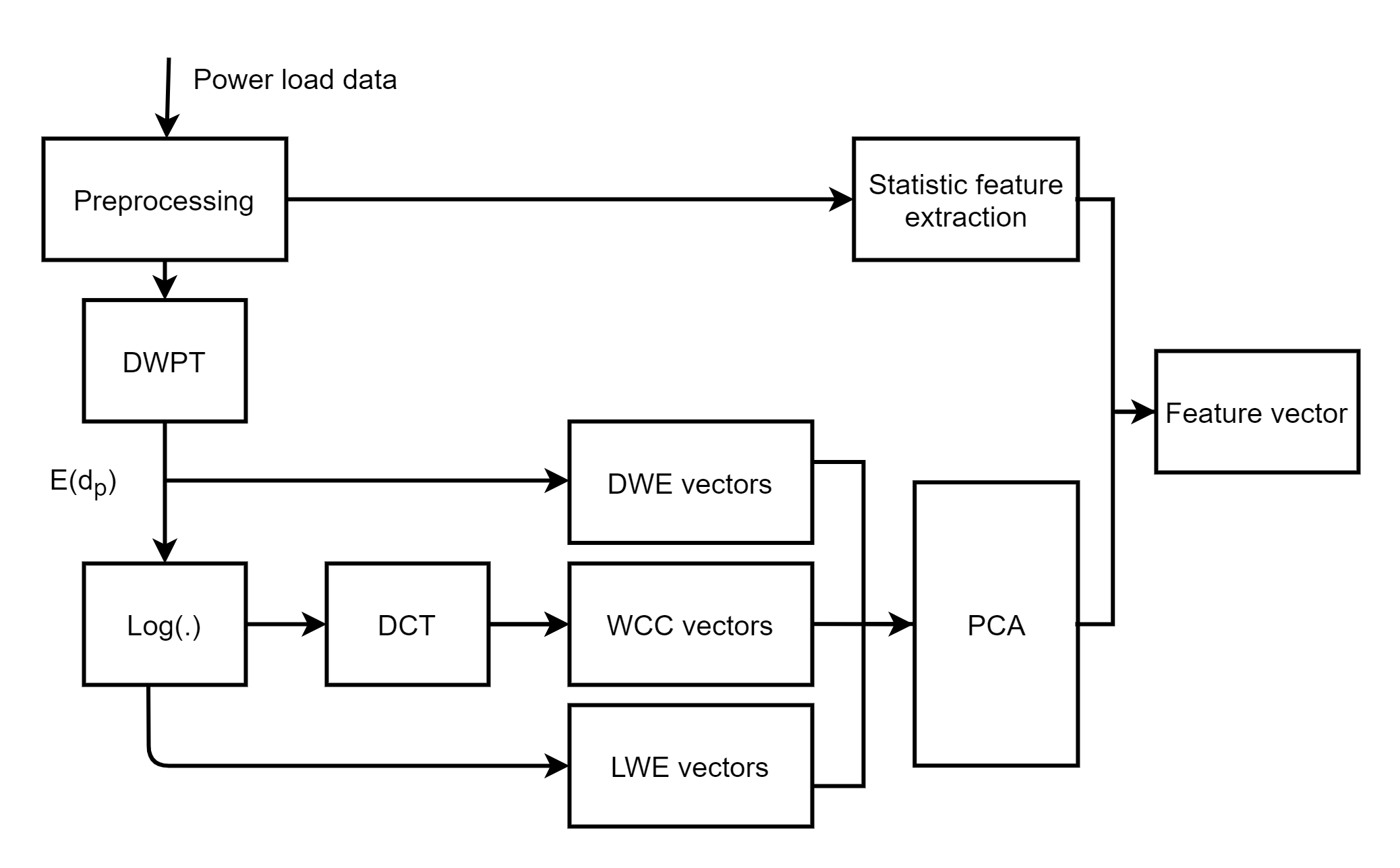}
    \caption{Feature Extraction Workflow}
    \label{Figure3}
\end{figure}

Consider a DWPT balanced tree with total ${L}$ levels of decomposition, the DWE value of a specific set of wavelet coefficient at level ${j}$, denoted by ${DWE}\left({d}_{{j}}\right)$, with ${N}_{j}$ number of detailed coefficients within the level, is calculated as:
\begin{equation}
    {DWE}\left({d}_{{j}}\right)=\frac{1}{{E}} \sum_{{n}=1}^{{N}_{{j}}}\Vert {~d}_{{j}}[{n}] \Vert_2 ^{2}, {1} \leq {j} \leq {L}
\end{equation}
where the ${l}_2$-norm of each wavelet coefficient ${d}_{{j}}$ is scaled to the total energy ${E}$ of all levels. The LWE is then calculated by applying ${log}_{10}$ to DWE feature vectors in order to achieve decorrelation of the energy values between different levels, which is defined as:
\begin{equation}
    {LWE}\left({d}_{{j}}\right)=\log \left(\frac{1}{{E}} \sum_{{n}=1}^{{N}_{{j}}}\Vert{~d}_{{j}}[{n}]\Vert_2^{2}\right)
\end{equation}

Based on the result of LWE, we further calculate the WCC feature vectors by applying DCT:
\begin{equation}
    {WCC}\left({d}_{{j}}\right)={DCT}\left[\log \left(\frac{1}{{E}} \sum_{{n}=1}^{{N}_{j}}\Vert{~d}_{{j}}[{n}]\Vert_2^{2}\right)\right]
\end{equation}

After the derivative of WCC, we combine DWE, LWE, and WCC feature vectors into one feature vector and apply Principle Component Analysis (PCA) to reduce the dimensionality of the feature space further.

\subsubsection{Other statistical features}
We further introduce several statistical features directly extracted from the time domain to represent time series data comprehensively. 
\begin{itemize}
    \item Seasonal and trend indicators: According to \cite{cleveland1990stl}, seasonal and trend decomposition based on loss (STL) suggests that any time series $X_{t}=\left\{x_{1}, x_{2}, \cdots, x_{N}\right\}$ can be decomposed in to three additive components: $X_{t}=T_{t}+S_{t}+E_{t}$, where $T_{t}$ is the tendency component, $S_{T}$ is the seasonal component, while $E_{t}$ stands for residual component. To measure the trend and periodical behavior of the original series, we define the following indices respectively:
    \begin{equation}
    \begin{aligned}
        s_{{deg}}=1-\frac{\operatorname{var}\left(E_{t}\right)}{\operatorname{var}\left(X_{t}-T_{t}\right)} \\
        t_{{deg}}=1-\frac{\operatorname{var}\left(E_{t}\right)}{\operatorname{var}\left(X_{t}-S_{t}\right)}
    \end{aligned}
    \end{equation}
    \item Skewness: The skewness represents the heavy tail (asymmetric) phenomenon of a probability distribution. For a normal distribution, the skewness is equal to 0. In this perspective, skewness can be used as a measure of non-Gaussian property. The skewness of the random variable $X$ is defined as:
    \begin{equation}
        {skew}(X)=E\left[\left(\frac{X-\mu}{\sigma}\right)^{3}\right]
    \end{equation}
    \item Sample entropy: As stated in \cite{richman2000physiological}, sample entropy is a metric measuring the non linearity of time series. For a time series $X_{t}=\left\{x_{1}, x_{2}, \cdots, x_{N}\right\}$, we sample the original series into ${N-m+1}$ segments with a template vector of length ${m}$ defined as:
    \begin{equation}
        X_{m}(i)=\left\{x_{i}, x_{i+1}, \cdots, x_{i+m-1}\right\}, {1} \leq {i} \leq {N-m+1}
    \end{equation}
    We further compute the distance between segments $i$, $j$, $i \neq j$ as:
    \begin{equation}
        d\left[X_{\mathrm{m}}(i), X_{\mathrm{m}}(j)\right]=\max\limits_{{k=0,...,m-1}}\left\|x_{i+k}-x_{j+k}\right\|
    \end{equation}
    
    For a given threshold ${r}$, we count the number of segments pairs with $d\left[X_{m}(i), X_{m}(j)\right]<r$ as ${N}_{m}$, and the number of pairs with $d\left[X_{m+1}(i), X_{m+1}(j)\right]<r$ as ${N}_{m+1}$. For finite number ${N}$, the sample entropy is then calculated as:
    \begin{equation}
        {SampEn}=-\ln \frac{N_m}{N_{m+1}}
    \end{equation}
    Considering the extreme few-shot scenario (i.e., 12 shots), where the total number of segments may be limited for large $m$, we directly set $m = 2$ and $r = 0.2 \times std(X_{t})$.
    
    \item Hurst exponent: As a coefficient describing autocorrelation, Hurst exponent is a nonlinear metric for long-term dependency of a sequence\cite{weron2002estimating}. We denote the standardized series as: 
    \begin{equation}
      X^{'}_{t}=\frac{{X_t -mean(X_t) }}{{std(X_t)}},  
    \end{equation}
    and calculate the ${i^{th}}$ element of the cumulative sum sequence ${Y_t}$ as: 
    \begin{equation}
        y_i= {\mathop{ \sum }\nolimits_{{k=1}}^{{i}}{x^{'}_{{k}}}}
    \end{equation}
    where $x^{'}_{{k}}$ is the ${k^{th}}$ element of $X^{'}_{t}$. The Hurst exponent is then calculated as:
    \begin{equation}
        K=\frac{2}{N} \log (\max (Y(t))-\min (Y(t)))
    \end{equation}
\end{itemize}

\subsubsection{Ensemble clustering}:Note that clustering models such as K-means and Gaussian mixture model (GMM-EM) have high sensitivity to initial values. The clustering result may vary when applying different models with distinct initial values on the same sample space. Therefore, in order to acquire stable clustering results, we follow a clustering ensemble method based on the hypergraph algorithm introduced in \cite{strehl2002cluster}, namely, clustering-based similarity partition algorithm (CSPA). To ensemble the clustering results generated by multiple models and attempts, binary similarity matrices $H$ are formulated to capture the pairwise similarity between clustering results, while the co-association matrix is computed as $S\ =\ HH\mathop{{}}\nolimits^{{T}}$. Then a hypergraph is generated based on a co-association matrix, where vertex represents time series sample, and an edge represents the similarity between objects. Finally, METIS \cite{karypismetis} algorithm based on graph theory is used to obtain the final clustering results. The structure of ensemble clustering is visualized in Fig. 4.
\begin{figure}[htbp]
    \centering
    \includegraphics[width=0.5\linewidth,scale=1.00]{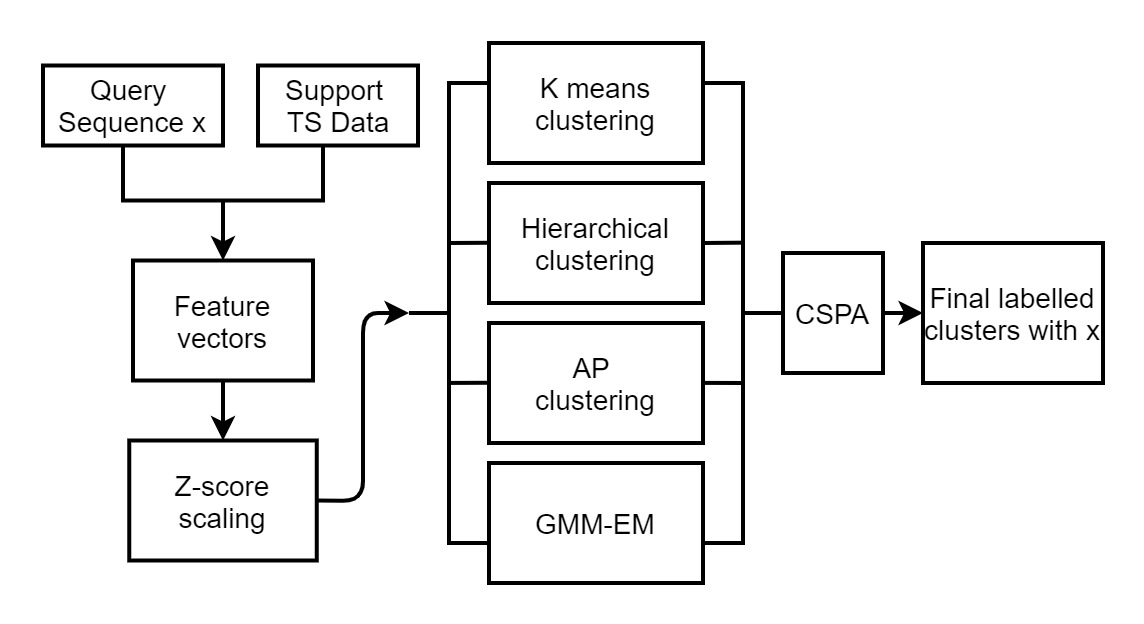}
    \caption{Ensemble Clustering}
    \label{Figure4}
\end{figure}

\subsection{LSTM-based Few-shot Forecasting}
\subsubsection{Wavelet denoising}
To achieve FSL, we attempt to acquire prior knowledge about the characteristics of few-shot series, to generate a pre-trained model. By averaging all historical data from the clustering results, we obtain one sample series for each cluster, namely prototype series. The model then obtains a set of denoised prototype series and few-shot time series by performing DWT with a hard threshold. The continuous wavelet transform (CWT) is given by: 
\begin{equation}
    H(x)=\frac{1}{|\sqrt{\zeta}|} \int x(t) \cdot \overline{\psi} \left(\frac{t-\tau}{\zeta}\right) dt
\end{equation}
where signal $x(t)$ has a wavelet transform result as a function of time $(t)$. $\psi$ is a mother wavelet continuous in both time and frequency domain, and $\overline{\psi}$ represents the complex conjugate of $\psi$. $\zeta$ is the scale parameter. $\tau$ is the transitional parameter. The DWT of the signal $x(t)$ is calculated by passing it through high and low pass filters. The decomposition of DWT is chosen to stop when the coefficients in the output are corrupted by edge effects caused by signal extension, where $l_{x}$ is the length of signal and $l_{f}$ is the length of the filter.
\begin{equation}
    level =\left \lfloor    log_{2}\left(\frac{l_{x}}{l_{f}}\right) \right \rfloor 
\end{equation}
The hard threshold is implemented with $T$ denoted as the given threshold.
\begin{equation}
    \operatorname{\rho_{T}}(x)=\left\{\begin{array}{cc}
x+T & x \leq-T \\
0 & |x| \leq T, \\
x-T & x \geq T
\end{array}\right.
\end{equation}
%\begin{figure}[htbp]
    %\centering
    %\includegraphics[width=\linewidth,scale=1.00]{DN.png}
    %\caption{DWT denoising with hard threshold}
    %\label{Figure4}
%\end{figure}

\begin{figure}[htbp]
    \centering
    \includegraphics[width=0.5\linewidth,scale=1.00]{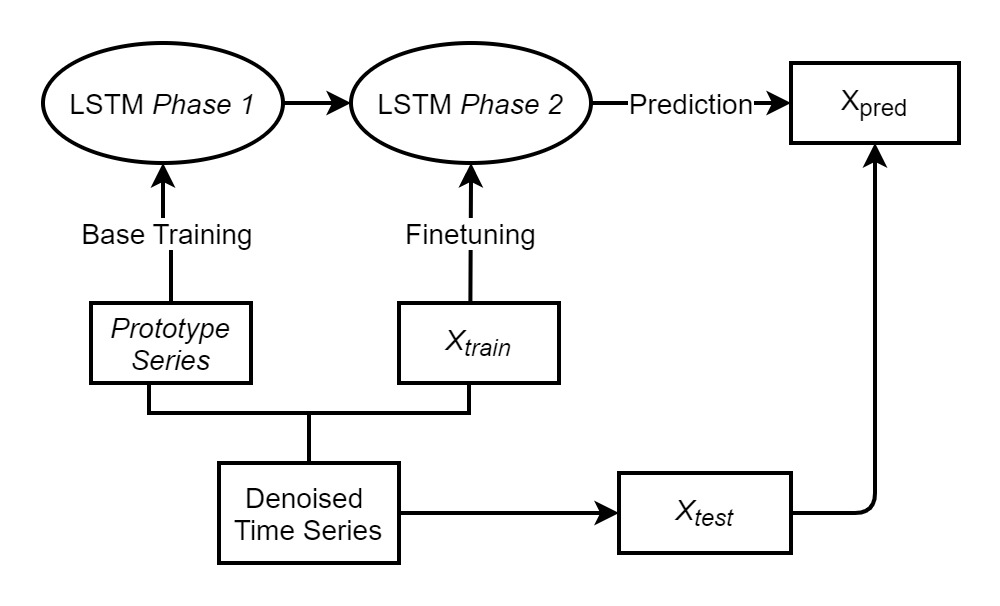}
    \caption{Two-Phase LSTM}
    \label{Figure5}
\end{figure}

\subsubsection{Two-phase LSTM}
The model is designed to make full use of prior knowledge extracted from unsupervised ensemble clustering. Allocated in the same cluster $c_{a}$, a set of historical data $X^{S_{1}},X^{S_{2}}...,X^{S_{n}}$ with abundant data points and  few-shot time series $X^{Q} = (x_{1},x_{2}...,x_{m})$ share similar features that can be learnt as prior knowledge by two-phase LSTM (structure shown in Fig. 5).
\begin{itemize}
    \item $Phase$ 1: The prototype series of historical data in  $c_{a}$, $X^{c}$, is used to train the basic LSTM's network weights to $\theta_{0}$, where the network possesses the ability to fast adaption to forecasting tasks for the new user in phase 2.
    \item $Phase$ 2: The few-shot time series $X^{Q}$ are split into $(X^{Q}_{train},X^{Q}_{test})$, where $\left |X^{Q}_{train} \right| \ll \min_{i}\left | X^{S_{i}} \right |$; $X^{Q}_{train}$ fine-tunes $\theta_{0}$ to $\theta_{1}$; $X^{Q}_{test}$ is used in the testing of few-shot task.
\end{itemize}

\section{Setup for Case Study}
In this section, we introduce the performance metrics and overview the datasets for our case study.

\subsection{FSL Task Formulation}
The experiment tries to discover the performance of the proposed FSL under different levels of data shortage, namely trained with 12, 24, 48, 96, 192 shots of training data. For few-shot time series $X^{Q}$ in $k$-shot learning scenario, $(x_{1},.,x_{k})$ is used in unsupervised clustering together with historical data. In two-phase LSTM fine-tuning, the prototype series of clustering results supports the base training of the LSTM model. The $k$-shot data is used in the second phase to fine-tune LSTM.
A fixed section of $X^{Q}$ with length 72, $(x_{k+1},.,x_{k+72})$ is used as ground truth in testing.

% \usepackage{multirow}

% \usepackage{multirow}

\subsection{Metrics}
Root Mean Square Error (RMSE) is one of the most used performance evaluation factors for forecasting or analyzing time series. For $n$ testing data, denote $p_x$ as the ground truth and $\hat{p_{x}}$ as the corresponding forecast value, such that $x = 1$ to $N$, the RMSE is given as,
 \begin{equation}
     RMSE = \frac{1}{n}  {\sum_{x=1}^{n}}  \sqrt{(p_x-\hat{p_x})^{2}}
 \end{equation}
In our FSL settings, to describe the model's overall performance of multiple predictions on different time series in $c_{a}$, Mean Root Mean Square Error (MRMSE) is introduced. For $M$ time series, the MRMSE is given as,
\begin{equation}
    MRMSE = \frac{1}{Mn}{\sum_{i=1}^{M}} {\sum_{x=1}^{n}}  \sqrt{(p_{ix}-\hat{p_{ix}})^{2}}, i \in c_{a} 
\end{equation}

To eliminate outliers in our result, we cover the $95\%$ confidence interval by adding or subtracting the MRMSE by two standard deviations and deleting values outside the interval. The mean and standard deviation of the remaining RMSE are then recalculated, and we use ${MRMSE} \pm {std}(RMSE) $ as our final metric to represent forecasting performance. 

\subsection{Experimental Setup}
Firstly, an ablation experiment on the two real-world datasets is conducted to compare our model with traditional LSTM. Then, our model is applied to the synthetic dataset to verify a theoretical lower bound of shots. Lastly, we perform a sensitivity analysis on the proposed model based on the experiment, which investigates the influence of cluster compactness on forecast accuracy. 

The clustering model is trained on an ensemble clustering model consisting of K-means, GMM-EM, hierarchy clustering, and affinity propagation, where the maximum level of DWPT ${L} = 5$. Hyperparameters used in LSTM model training and fine-tuning include batch size (72), initial learning rate (0.001), training steps (130 for pretraining and 70 for fine-tuning), and optimizer (Adam). All the experiments are performed on a Linux server with an Intel Xeon E5-2620@2.10 GHz and 128GB of RAM.

\subsection{UMass Smart Dataset}

UMass Smart Dataset (2017 release)\cite{barker2012smart} includes minute-level electricity usage data from more than 400 anonymous homes.
The dataset is sliced to have the time span from January 1, 2016 to March 10, 2016. During this period, 114 homes' records are available. The granularity is set to be 20 minutes, 1 hour, 2 hours by averaging over data:
\begin{equation}
    y_{a}[m]=\frac{1}{k} \sum_{i=mk}^{mk+k-1} m[i]
\end{equation}
The FSL-LSTM is trained with 12, 24, 48, 96, 192 shots. A fixed section of $X^{Q}$ with a length of 72 is used for testing. Fig. 6 visualizes the UMass electricity load.

%\begin{figure}[htbp]
%\centering
%\begin{minipage}[t]{0.24\textwidth}Caption
%\centering
%\includegraphics[width=5cm]{pecan12shot.jpg}
%\caption{World Map}
%\end{minipage}
%\begin{minipage}[t]{0.24\textwidth}
%\centering
%\includegraphics[width=5cm]{apt12shot.jpg}
%\caption{Concrete and Constructions}
%\end{minipage}
%\end{figure}
    %\centering
    %\includegraphics[width=0.75\linewidth,scale=0.50]{apt12shot.jpg}
    %\caption{Case 1, FSL-LSTM 12-shot prediction result with ground %truth }
    %\label{}
%\end{figure}
%\begin{figure}[htbp]
    %\centering
    %\includegraphics[width=0.75\linewidth,scale=0.50]{Umas_RMSE.jpg}
    %\caption{Case 1, Performance Comparison of FSL-LSTM, LSTM, ARIMA %in different few shot scenario}
    %\label{}
%\end{figure}
\begin{figure}[h]
    \centering
    \includegraphics[width=0.5\linewidth,scale=0.70]{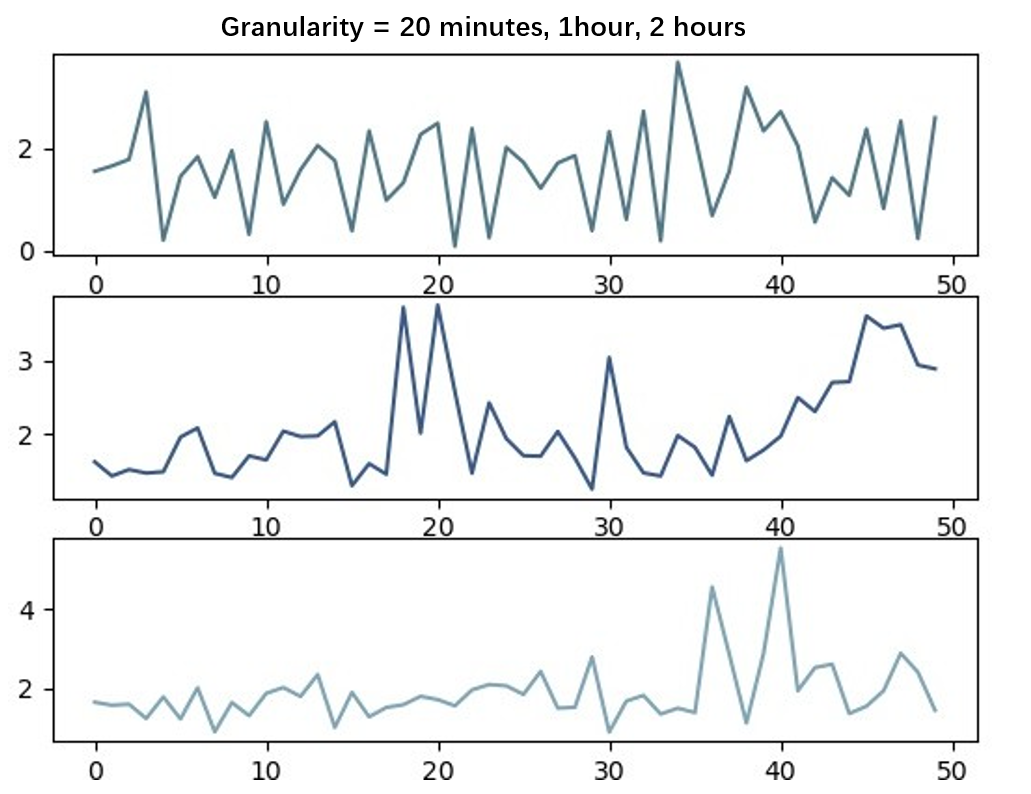}
    \caption{UMass Smart Dataset with Different Granularity}
    \label{Figure6}
\end{figure}

\begin{table*}[htbp]
%\makeatleater
%\patchcmd{
%\makecaption{Table 1, Comparison of Prediction Accuracy on UMass Smart Dataset in k-shot Learning Scenario}}
\caption{FSL-LSTM Prediction Accuracy (MRMSE) on UMass Smart Dataset}
%\makeatleater
\centering
\begin{tabular}{cl|lllll}
\hline
\multicolumn{2}{c}{Dataset}                                           & \multicolumn{5}{c}{Umass}                            \\ \hline
\multicolumn{1}{l}{Granularity} & \multicolumn{1}{l|}{Methods}        & 12shot   & 24shot   & 48shot   & 96shot   & 192shot  \\ \hline
\multirow{2}{*}{20 minutes}  &{FSL-LSTM(Ours)}  & 0.883$\pm0.317$ &0.999$\pm0.366$ &0.959$\pm0.296$ &0.931$\pm0.336$ &1.004$\pm0.377$  \\ 
                          &{LSTM}  & 1.177$\pm0.483$  &1.240$\pm0.443$& 1.499$\pm0.211$& 1.096$\pm0.340$ & 1.004$\pm0.339$  \\ \hline
                          
\multirow{2}{*}{1 hour}   &{FSL-LSTM(Ours)} & 0.693$\pm0.306$ &0.738$\pm0.314$ &0.423$\pm0.228$ &0.551$\pm0.330$ &0.317$\pm0.123$ \\
                         &{LSTM}           & 0.748$\pm0.400$ & 0.844$\pm0.404$ & 0.510$\pm0.155$ &0.437$\pm0.205$ &0.434$\pm0.179$ \\ \hline
                         
\multirow{2}{*}{2 hours}   &{FSL-LSTM(Ours)}& 0.528$\pm 0.233$& 0.339$\pm0.179$& 0.347$\pm0.146$& 0.308$\pm0.086$& 0.308$\pm0.197$ \\
                          &{LSTM}           & 0.695$\pm 0.352$& 0.283$\pm0.161$& 0.754$\pm  0.277$ & 0.335$\pm0.127$& 0.321$\pm  0.144$\\ \hline
                
\end{tabular}

\end{table*}
\begin{table*}[htbp]
\centering
%\makeatleater
%\patchcmd{
%\makecaption{Table 2, Comparison of Prediction Accuracy on Pecan Street Dataset in k-shot Learning Scenario}}
%\makeatleater
\caption{FSL-LSTM Prediction Accuracy (MRMSE) on Pecan Street Dataset}

\begin{tabular}{cl|lllll}
\hline
\multicolumn{2}{c}{Dataset}                      & \multicolumn{5}{c}{Pecan Street}                                                       \\ \hline
\multicolumn{1}{l}{Granularity} & Methods        & 12shot          & 24shot          & 48shot          & 96shot          & 192shot         \\ \hline
\multirow{2}{*}{20 minutes}      & FSL-LSTM(Ours) & 0.388$\pm0.190$ & 0.308$\pm0.161$ & 0.324$\pm0.185$ & 0.218$\pm0.104$ & 0.312$\pm0.164$ \\
                                & LSTM           & 0.662$\pm0.294$ & 0.675$\pm0.420$    & 0.396$\pm0.174$ & 0.314$\pm0.141$ & 0.338$\pm0.194$ \\ \hline
\multirow{2}{*}{1 hour}         & FSL-LSTM(Ours) & 0.418$\pm0.187$ & 0.363$\pm0.154$ & 0.512$\pm0.188$ & 0.515$\pm0.198$ & 0.466$\pm0.314$ \\
                                & LSTM           & 0.624$\pm0.287$ & 0.521$\pm0.230$ & 0.670$\pm0.212$ & 0.543$\pm0.214$ & 0.593$\pm0.282$ \\ \hline
\multirow{2}{*}{2 hours}         & FSL-LSTM(Ours) & 0.422$\pm0.226$ & 0.533$\pm0.260$ & 0.337$\pm0.217$ & 0.416$\pm0.203$ & 0.382$\pm0.143$ \\
                                & LSTM           & 0.536$\pm0.316$ & 0.465$\pm0.263$ & 0.617$\pm0.283$ & 0.527$\pm0.296$ & 0.394$\pm0.129$ \\ \hline
\end{tabular}

\end{table*}
\begin{figure}[htbp]
    \centering
    \includegraphics[width=0.5\linewidth,scale=0.70]{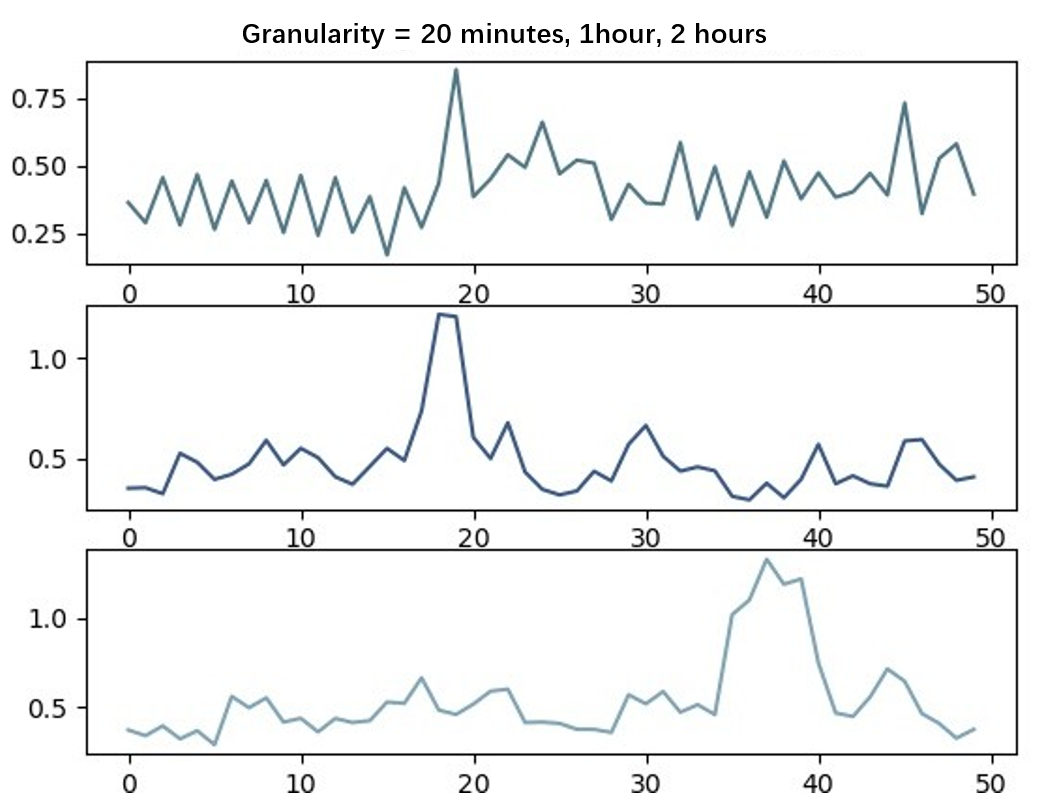}
    \caption{Pecan Street Dataset with Different Granularity}
    \label{Figure7}
\end{figure}
\subsection{Pecan Street Texas Dataset }
Pecan Street Dataport \cite{pecan}
includes minute-level electricity usage data from 310 units in Texas.
The dataset is sliced to have a time interval from January 1, 2016 to March 10, 2016. The granularity is set to 20 minutes, 1 hour, 2 hours by averaging over data. The FSL-LSTM is trained with 12, 24, 48, 96, 192 shots. A fixed section of $X^{Q}$ is used for testing. Fig. 7 shows the visualization of Pecan Street electricity load.

\subsection{Synthetic Dataset}
Since users' real-world power load data is not always based on a 24-hour cycle, we designed a synthetic dataset consisting of sinusoidal waves where Gaussian noise is constructed to explore the influence of data cycle and training length on the model performance. The periods of time series are set to be  10, 15, and 20 sample points.

\section{Numerical Results and Analysis}

The Pecan Street and Umass dataset experiment suggest that FSL-LSTM outperforms traditional LSTM in most FSL scenarios. The detailed MRMSE results in Table \RNum{1} and \RNum{2} show significant improvements in precision and variance for forecasting 20 minutes, 1 hour, and 2 hours-level energy load in FSL. As shot length increases, the proposed method is followed more closely by traditional LSTM.
\subsection{Influence of $k$ Shot}
In order to present the significant advantages of our proposed method over the traditional training method under extreme data shortage of new users, we consider $k$ = {12, 24, 48, 96, 192} for $X^{Q}_{train}$ and measure the overall performance using MRMSE. The results are shown in Fig. 8 and Fig. 9.

As an FSL forecasting model, fine-tuning is the most crucial stage relating to forecasting accuracy. One intuitive assumption is that the model's forecasting performance will improve with the length of the fine-tuning samples (i.e., the length of the few-shot samples). For our model, this assumption coincides with the global observation, where the MRMSE decreases with the increase of $k$. Moreover, FSL-LSTM significantly outperforms the traditional LSTM in terms of forecasting accuracy for general cases. For example, when $k$ = 12, we observe a considerable gap of MRMSE between the proposed method and traditional LSTM in both real-world datasets, where the MRMSE of FSL-LSTM is significantly lower than the traditional LSTM. The excellent performance of FSL-LSTM is unexpected as given just 12 data points with granularity ranges from 20 minutes to 2 hours, which even fail to cover an entire period of usage pattern. The global observation of relatively low MRMSE using FSL-LSTM compared to traditional LSTM validates the theoretical advantages of our proposed method over the traditional one.

\begin{figure}[htbp]
    \centering
    \includegraphics[width=0.5\linewidth,scale=0.70]{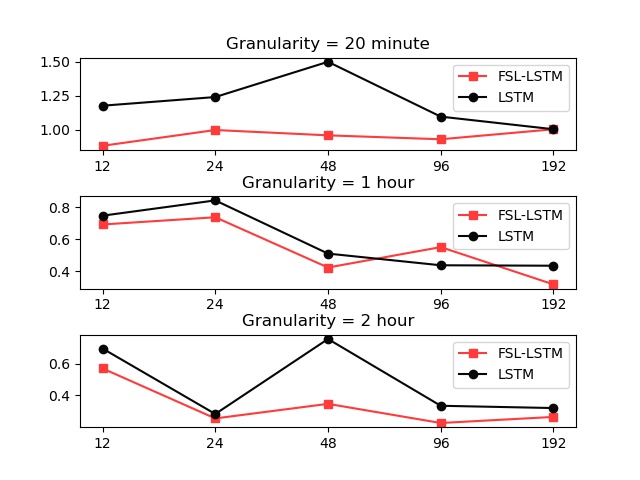}
    \caption{Case 1: MRMSE on Umass Dataset}
    \label{figure9}
\end{figure}

\begin{figure}[htbp]
    \centering
    \includegraphics[width=0.5\linewidth,scale=0.70]{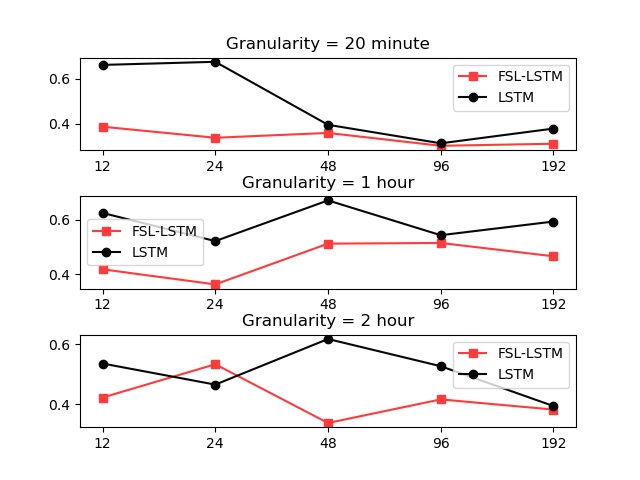}
    \caption{Case 2: MRMSE on Pecan Street Dataset}
    \label{figure10}
\end{figure}

\begin{table*}[htbp]
\centering
%\makeatleater
%\patchcmd{
%\makecaption{Table 3, Comparison of FSL-LSTM Prediction Accuracy with}
%}
%\makeatleater
\caption{FSL-LSTM Prediction Accuracy (MRMSE) with Different Clustering Models}
\begin{tabular}{l|llllll}
\hline
$k$-shot             & 12               & 24              & 48              & 96                                              & 192             & S-score \\ \hline
K-means               & 0.534$\pm0.371$ & 0.424$\pm0.403$ & 0.350$\pm0.359$ & 0.345$\pm0.314$                                 & 0.336$\pm0.282$ & 0.1385  \\ \hline
Agglomerative        & 0.509$\pm0.175$  & 0.414$\pm0.307$ & 0.338$\pm0.274$ & 0.326$\pm0.178$                                 & 0.315$\pm0.286$ & 0.3843  \\ \hline
GMM-EM               & 0.522$\pm0.427$  & 0.426$\pm0.311$ & 0.344$\pm0.328$ & 0.347$\pm0.266$ & 0.354$\pm0.185$ & 0.1230   \\ \hline
Affinity Propagation & 0.519$\pm0.293$     & 0.426$\pm0.276$ & 0.345$\pm0.326$ & 0.345$\pm0.314$                                 & 0.345$\pm0.271$ & 0.2138  \\ \hline
Ensemble          & 0.500$\pm0.242$     & 0.407$\pm0.143$ & 0.314$\pm0.172$ & 0.306$\pm0.220$                                 & 0.316$\pm0.197$ & 0.3622  \\ \hline
\end{tabular}

\end{table*}

\subsection{Influence of Granularity}
Since we extracted only short segments of sequence from the historical dataset in order to match with the length and the point in time of few-shot time series, when the length of the few-shot series fails to cover a whole period, namely ${T}$, of the ground truth series, clustering results at first stage does not necessarily guarantee the following trends are similar to each other. Theoretically, to avoid mislabeling, the length of few-shot series for fine-tuning, denoted by ${N}$, is expected to be ${N}\geq T/M$ for a fixed granularity ${M}$. This lower bound is particularly phenomenal in our synthetic dataset while not violating the observation in the real-world datasets.

When granularity is small, the ideal length of few-shot samples that yield acceptable MRMSE is significantly larger than that of large granularity. Furthermore, the granularity and few-shot length pairs reach the most benign model performance when their products fully contain one or multiple periods of the historical dataset. This phenomenon is much more significant on our synthetic dataset. As shown in Fig. 10, the model reaches the lowest MRMSE when ${N} = PT/M$, where ${P}$ denotes any positive integer. The MRMSE then remains relatively steady after ${N}$ reaching the threshold, which means that our theoretical assumptions do not violate empirical observation.

\begin{figure}[htbp]
    \centering
    \includegraphics[width=0.5\linewidth,scale=0.70]{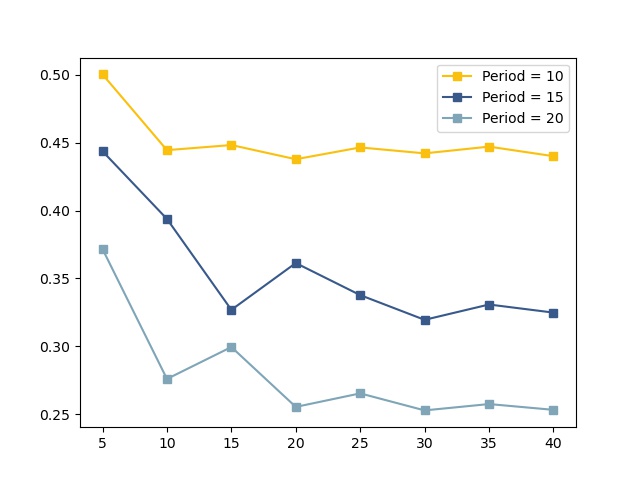}
    \caption{Case 3: MRMSE on Synthetic Dataset}
    \label{figure11}
\end{figure}

\subsection{Influence of Cluster Compactness}
As an FSL forecasting model, the prediction accuracy of the fine-tuned model depends on the quality of the prior knowledge. One rational intuition is that the compactness of clustering results is positively correlated with MRMSE. We conduct single factor sensitivity analysis by changing different clustering models on the UMass Smart dataset with 1-hour granularity to investigate the hypothesis. In order to quantify the compactness of clusters, the Silhouette score (S-score) is introduced. The results are shown in Fig. 11. 

\begin{figure}[htbp]
    \centering
    \includegraphics[width=0.5\linewidth,scale=0.70]{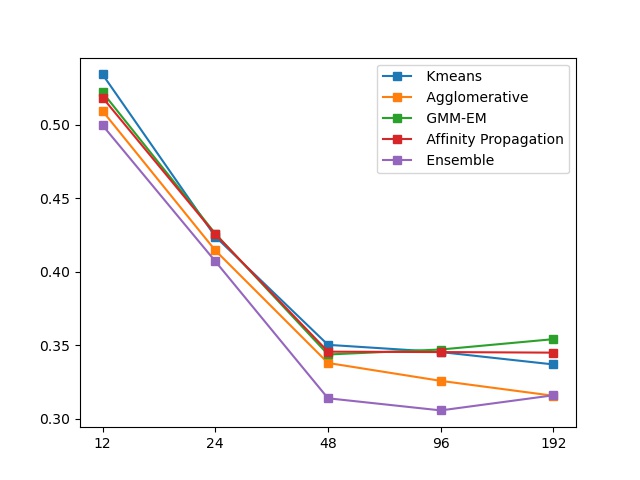}
    \caption{Case 4: MRMSE for Different Clustering Models}
    \label{figure12}
\end{figure}

Table \RNum{3} suggests that the S-score of ensemble clustering is higher than those of traditional clustering models due to the elimination of some edge samples. Moreover, the standard deviation of RMSE has a negative correlation with S-score. This means that the larger the S-score, the more likely the denoised prototype can capture most of the local features inside the cluster. In addition, MRMSE reduces slightly when S-score improves. However, the difference of MRMSE for a static shot between different clustering models is not significant.

\section{Conclusion and Future work}
Quickly adapting to time series forecasting tasks with limited customized samples is essential for electricity load forecasting and other practical applications. We contribute to this field by proposing the FSL time series forecasting based on LSTM. The proposed method leverages the existing power load records through ensemble clustering to gather an ability to solve few-shot forecasting tasks on previously unseen time series efficiently. Numerous studies suggest that the proposed method can vastly outperform its baseline on two electricity load datasets. Moreover, we empirically interpret FSL-LSTM's performance from two aspects, $k$-shot setting and granularity of data.
 
In the future, it would be interesting to explore more sophisticated few-shot learning techniques such as \cite{santoro2016meta,articalMAML} for load forecasting. Besides, by combining FSL with incremental learning \cite{yoon2020xtarnet}, a robust AI blueprint can be provided to the power grid system, such that models can be swiftly generated through FSL when the data scale is small and be fine-tuned locally as data scale increases. 

\bibliographystyle{IEEEtran}
\bibliography{references}
%\bibliography{reference}

%\printbibliography

\end{document}